# Associative Memories Based on Multiple-Valued Sparse Clustered Networks

Hooman Jarollahi*, Naoya Onizawa†, Takahiro Hanyu† and Warren J. Gross*
*Department of Electrical and Computer Engineering, McGill University, Montreal, Quebec H3A 0E9
† Research Institute of Electrical Communication, Tohoku University, Sendai, Japan
Email: hooman.jarollahi@mail.mcgill.ca, {onizawa, hanyu}@ngc.riec.tohoku.ac.jp, warren.gross@mcgill.ca

*Abstract*—Associative memories are structures that store data patterns and retrieve them given partial inputs. Sparse Clustered Networks (SCNs) are recently-introduced binary-weighted associative memories that significantly improve the storage and retrieval capabilities over the prior state-of-the art. However, deleting or updating the data patterns result in a significant increase in the data retrieval error probability. In this paper, we propose an algorithm to address this problem by incorporating multiple-valued weights for the interconnections used in the network. The proposed algorithm lowers the error rate by an order of magnitude for our sample network with 60% deleted contents. We then investigate the advantages of the proposed algorithm for hardware implementations.

## I. INTRODUCTION

Data storage and retrieval methodologies in associative memories are different from the widely-known indexed memories in which the data is written and accessed using explicit addresses. In associative memories, only associations between parts of data patterns are stored in a way that data can later be accessed by presenting a partial input pattern. Associative memories are suitable for applications such as database engines, data mining and implementation of sets [1]–[5], where the relevant data is searched using partial inputs. It is also possible to construct the function of an associative memory using indexed memories by performing exhaustive search operations on partial input patterns with the cost of large processing delays.

Classical associative memories implemented with Hopfield Neural Networks (HNN) [1] store input patterns (messages) in a network consisting of nodes (neurons) and connections between them, where the index of each data bit corresponds to that of a node. Integer-weights for the connections are then computed between the nodes such that each of the nodes has interconnections with all adjacent nodes. The decoding algorithm for HNNs retrieves messages from partial inputs using the stored integer weights. A drawback of HNNs is that the number of nodes in the network must be equal to the length of each message which, due to a fixed available memory capacity, limits the diversity of the stored messages - the number of different messages the network can store. Furthermore, the network is not efficient in terms of memory usage since the ratio between the number of bits stored to that used approaches zero as the network grows to increase its storage capacity. These drawbacks were addressed in Sparse Clustered Networks (SCNs) introduced in [2] [6], where the interconnections are binary and the nodes are clustered in such a way that a significantly larger diversity and memory efficiency are achieved compared to those of an HNN. The algorithms and architectures of various types of SCNs were proposed from a hardware design perspective in [3], [4], [7], where significant speed-up was achieved in hardware compared to its software counterpart. A drawback of the conventional SCNs is that due to the use of binary-weighted interconnections, deleting or updating messages causes the removal of the shared connections used by other stored messages. As we will show in this paper, the lost connections result in a significant increase in the Message Error Rate (MER) – the ratio between the average number of correctly retrieved messages to the total number of partial input messages selected from the previously stored ones. Some of the lost connections can be recovered due to the error-correcting capability of SCNs for which the degree or recovery depends on the selection of hardware architecture and the application. In this paper, we propose an algorithm based on a variation of the conventional SCN algorithm that incorporates Multiple-Valued (MV) interconnections. In the proposed algorithm (MV-SCN), deletions and updates of messages are achieved with a significant improvement in the MER compared to that of its binary-weighted counterpart. Section II reviews the concept of SCNs. In Section III, the proposed algorithm, addressing the MER-increase after deletion or updating messages, is presented. In Section IV, the behaviour of three hardware architectures based on MV-SCN under various conditions are analysed and compared against the previous approaches, where we show a problem with the first architecture and address it in the next two architectures. Two architectures are therefore recommended for use in the related applications: The first one optimized for aggressively-updated, low-MER applications, and the second, suitable for applications requiring a low hardware complexity and low-MER but more relaxed deletion or updating requirements. A trade-off analysis is then followed to elaborate the benefits of each architecture that depend on the application requirements. Both of the recommended architectures can be tuned for a desired MER with the cost of hardware resources. Section V concludes the paper.

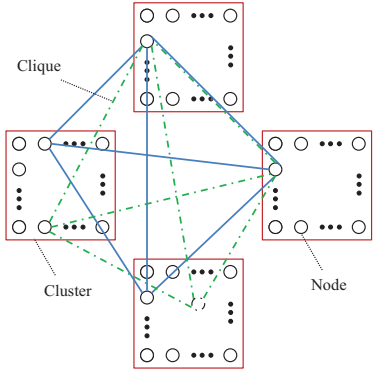

Fig. 1. Graphical representation of an SCN with 4 clusters storing two messages shown as cliques which share a connection that is removed in case of deleting one of the messages.

## II. REVIEW OF SPARSE CLUSTERED NETWORKS AS ASSOCIATIVE MEMORIES

The concept of SCNs, introduced as associative memories in [2], has been inspired by HNNs in which a message can be represented in a network containing nodes and weighted connections (links). As opposed to HNNs, the links are binary and the nodes are not fully interconnected. As shown in Fig. 1, an SCN network consists of $n$ nodes divided into $c$ clusters, with binary connections between each node to nodes in other clusters. No connections exist within a cluster between two nodes. The number of nodes per cluster does not necessarily have to be the same for each cluster. However, it is commonly chosen to be equal for all clusters. In this paper, we consider the number of nodes in a cluster to be equal to $l$ for all clusters. An input pattern (message), $m_i$ ($1 \leq i \leq M$), with a bit-length of $K$ bits is first divided into $c$ parts, $\kappa = K/c$ bits each. Then, each sub-message, $K_j$ ($1 \leq j \leq c$) is mapped to a node in its corresponding cluster through a process known as *Local Decoding* that activates a node, i.e., it sets the value of the node to '1' and the rest of the nodes to '0'. Therefore, in order to permit a one-to-one relationship between the any value of sub-message and the indices of the nodes of its corresponding cluster, the number of nodes, $l$, in a cluster must be equal to $2^{\kappa}$. If a sub-message contains erased bits, all possible values are computed in local decoding, and all corresponding nodes are activated. Therefore, if a sub-message is entirely erased, all of the nodes of its corresponding cluster are activated.

### A. Data Storage

In order to store a message in an SCN network, local decoding is first performed on each sub-message. After the nodes are activated, binary connections between the activated nodes (which are not inside the same cluster), are added to the network such that all activated nodes are fully interconnected. The interconnections between the activated nodes pertaining to a single message is known as a *clique* and is shown in Fig. 1. The binary links for all the stored cliques are then stored in a memory module as discussed in [4], [7] employing various hardware resources and design techniques.

### B. Data Retrieval

Once all the messages are stored through the binary connections, SCN can retrieve partial messages when they are presented as inputs. Local decoding is performed similar to local decoding in the learning process to activate nodes in each cluster. The next step, *global decoding* [2], resolves ambiguities amongst several nodes in a cluster by using the stored inter-cluster links. For each node a score is first computed according to two criteria: i) the number of connections that were previously learned, towards the other activated nodes in other clusters, and ii) whether or not the node has been activated as a result of local decoding. Then, the node or nodes that achieve the highest score in each cluster remain or become activated. The global decoding process is iterative until the activation of the nodes converges to an an unchanging state compared to that of a previous iteration. Due to the possibility of the presence of erased bits in an input, it is possible that after the local decoding process more than one node is activated in each cluster. This condition is known as *ambiguity* in reconstructing a unique clique. The ambiguities are then attempted to be *disambiguated* (removed) by iteratively performing the global decoding process until convergence occurs. An algorithm for global decoding is proposed and studied in [2], and is implemented in hardware in [3]. It first computes a score for each node in parallel with other nodes, and then uses a $max$-function to perform the winner-take-all operation. This structure has the capability to correct erroneous bits in the inputs to some degree, and is formalized as follows:

$$\forall i,j, s_{(i,j)} = \sum_{i'=1}^{c} \sum_{j'=1}^{l} w_{(i,j)(i',j')} v_{(i',j')} + \gamma v_{(i,j)}, \quad (1)$$

where $s_{(i,j)}$ is the score of the $j$-th node of the $i$-th cluster at the end of an iteration, $w_{(i,j)(i',j')}$ is the binary connection from the $j$-th node of the $i$-th cluster to the $j'$-th node of the $i'$-th cluster, $v_{(i',j')}$ is binary value of the $j$-th node of the $i$-th cluster, and $\gamma$ a factor known as the *memory effect*, and is a real number larger than zero.

The final step in global decoding is known as the *Winner-Take-All* rule, and is expressed as follows:

$$v^*_{(i,j)} \leftarrow \begin{cases} 1, & \text{if } s_{(i,j)} = s_{max} \\ & \text{and } s_{max} \geq \sigma \\ 0, & \text{otherwise} \end{cases}, \quad (2)$$

where $\sigma$ is a threshold value that can be adjusted to improve computational speed in scenarios that the range of the maximum scores can be pre-determined. $v^*$ is the updated value of the node after an iteration. In this paper $\sigma$ is assumed to be equal to $c$, and $\gamma$ is equal to 1.

Various hardware implementations techniques of the global decoding process and their advantages have been proposed and studied in [3], [4], [7]. The global decoding method shown in (1) and (2) are resource-hungry when implemented in hardware, mainly due to the presence of the $max$-functions, and $(c-1)l$-input adders for each node. This global decoding algorithm was later simplified in [4], [6] by realizing the fact that it is sufficient to activate a node in a cluster if there exists at least one connection per clusters from the activated nodes after the local decoding process. The hardware implementation of the latter algorithm significantly reduces the hardware complexity and improves the operating frequency as discussed in [4]. The reduced-complexity algorithm is formalized as:

$$v^*_{(i,j)} = \left( \bigwedge_{\substack{i'=1 \\ i' \neq i}}^{c} \bigvee_{j'=1}^{l} w_{(i,j)(i',j')} v_{(i',j')} \right) \bigwedge v_{(i,j)}, \quad (3)$$

where $\bigwedge_{\substack{i'=1 \\ i' \neq i}}^{c}$ performs a $(c-1)$-input logical AND operation, and $\bigvee_{j'=1}^{l}$ performs an $l$-input logical OR operation. This method can only be used for applications in which bit-erasures exist, and will not work for erroneous inputs.

The density of a network is defined as the ratio between the number of used connections by the stored messages to that of the total number of possible connections. The density is derived in [2] using the number of messages, $M$, the number of nodes per cluster, and regardless of the number of clusters as follows:

$$d = 1 - \left(1 - \frac{1}{l^2}\right)^M. \quad (4)$$

As the value of the density is increased, the probability of the existence of the number of shared connections between different cliques is increased due to the fixed number of nodes in the network. Since the connections are binary, and that shared connections are inevitable, deletion of the messages and updating them with new ones will result in the loss of connections which might be recovered to some extent in the retrieval process using (1) and (2).

## III. MULTIPLE-VALUED WEIGHT-CODED SCN

The learning process of the proposed algorithm using multiple-valued weighted connections is similar to that of the conventional algorithm in [2] with the difference that the weight value of a connection between two nodes may be increased by 1 when a new clique is added to the network. If the added clique includes a shared connection with another clique, especially at higher densities, the value of the weight of that connection is increased. Similarly, the weight value of a correspondence connection is decreased by 1 in case of deletion of a clique. Therefore deleting or updating messages in the proposed algorithm affect the values of the connection weights as follows:

$$w^*_{(i,j)(i',j')} = \begin{cases} w+1, & \text{if (6) is satisfied and} \\ & \text{adding messages} \\ w-1, & \text{if (6) is satisfied and} \\ & \text{removing messages} \\ w, & \text{otherwise} \end{cases} \quad (5)$$

$$\begin{cases} i \neq i' \\ \text{and } \exists m \in \{m_1...m_M\} \\ C(m)_{K_i} = j \text{ and } C(m)_{K_{i'}} = j' \\ w_{MIN} \leq w \leq w_{MAX} \end{cases} \quad (6)$$

where $C(m)_{K_i}$ is the index of the activated node through the local decoding corresponding to the $K_i$-th sub-message in message $m$. In this paper, $w_{MIN}$ is set to 0. The $w+1$ is used when adding messages, whereas $w-1$ is used when removing one. In this paper, an SCN network consisting of $c = 8$ clusters, and $n = 128$ nodes is realized as also used in [2]–[4], [7].

### A. Non-normalized Decoding

Since deleting messages results in the loss of connections in the binary-weighted SCNs, we propose using multiple-valued weights for the connections instead of the binary ones. As a first step, we will use integer weights with a fixed maximum value ($w_{MAX}$), and use (1) and (2) directly for the decoding process. We will refer to the variant of a the architecture in [3] using this algorithm as Architecture I.

The problem with Architecture I is that since the integer values of the weights are directly used in the inputs of adders, some connections can contribute more than the others in computing the score of a node. In a case that a sub-message is erased, the connections from a node corresponding to the sub-message are not considered in computing the score. As a result, a false node might be activated when its score is maximum even though it does not receive all the necessary connection from all other clusters, and simply because of the integer weighted connections contributing to the score. For example, let us assume that an activated node from local decoding ($n_{i,j}$) must receive 3 connections from other clusters to achieve a maximum score of 4 to remain activated. On the other hand, if one of the connections has a weight value, $w_{(i,j)(i',j')} = 3$, and the weight values of the other two connections are 0, $n_{i,j}$ will still achieve a score of 4. This leads to the false activation of $n_{i,j}$. Therefore, although the non-normalized decoding algorithm we just described helps keeping the connections after deleting messages, it also results in the activation of a higher number of ambiguous nodes compared to a case, in which the connections would equally contribute in computing the score of a node.

### B. Normalized Decoding

To tackle the problem with Architecture I, it is possible to normalize the values of the connection weights to binary values in the global decoding process and use the global decoding method in (1) as follows:

$$\forall i, j, s_{(i,j)} = \sum_{i'=1}^{c} \sum_{j'=1}^{l} \psi_{(i,j)(i',j')} v_{(i',j')} + \gamma v_{(i,j)}$$

$$\psi_{(i',j')(i,j)} = \begin{cases} 1, & \text{if } w_{(i,j)(i',j')} \geq 1 \\ 0, & \text{otherwise} \end{cases}, \quad (7)$$

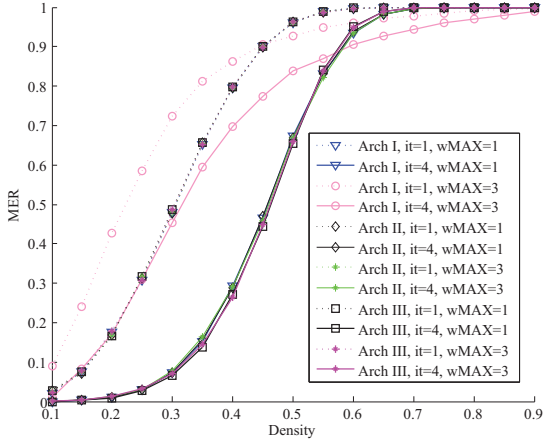

Fig. 2. The relationship between the network density and MER for $ce = 0.5$, no deletion or update, in different architectures and different values for the connection weights.

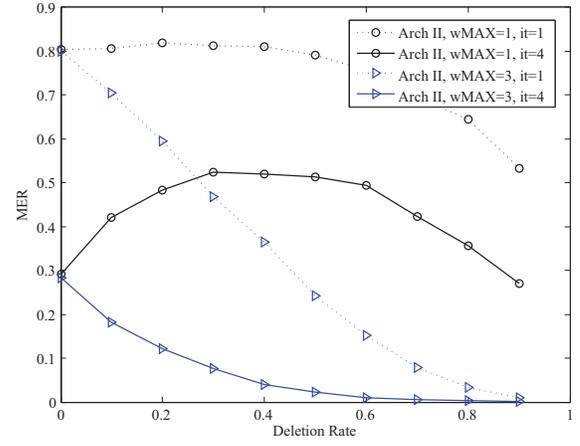

Fig. 3. The effect of increasing the deletion rate on MER in Architecture II for two values of the connection weights.

which is followed by (2). We refer to the method of exploiting multiple-valued weights in the learning and updating processes and normalization of the weights in the $max$-function-based decoding process as Architecture II.

Exploiting the multiple-valued weights in (3), a lower complexity architecture for the global decoder is achieved whose algorithm is expressed as:

$$v^*_{(i,j)} = \left( \bigwedge_{\substack{i'=1 \\ i' \neq i}}^{c} \bigvee_{j'=1}^{l} \psi_{(i,j)(i',j')} v_{(i',j')} \right) \bigwedge v_{(i,j)}, \quad (8)$$

where similar to (7), $\psi_{(i,j)(i',j')}$ is the normalized value of $w_{(i,j)(i',j')}$. We refer to the architecture achieved using (8) for the global decoding as Architecture III.

## IV. EVALUATION

The relationship between the network density, and the message error rate for Architectures I, II, and III, also showing the effects of the number of iterations ($it$) and increasing the maximum value for the connection weights from 1 to 3 (1 bit to 2 bits), is depicted in Fig. 2. It is assumed that 50% of the sub-messages are erased ($ce = 0.5$). It shows that the MER of Architecture I is increased when $w_{MAX}$ is increased from 1 to 3. This is because during the activation process of a node in the network, the non-normalized weights from shared connections contribute larger than non-shared connections in computing the score of that node. This causes some of the nodes to achieve maximum scores (and thus larger number of ambiguities) which would not have occurred if normalized weights were used as shared connections would not cause larger contributions in computing the scores. Architectures II and III show similar error performance when no deletion or update occurs for both binary or multiple-valued weights for the connections.

In order to investigate the benefit of multiple-valued connection weights in Architecture II, we first select a target density (0.4 for example) that results in a desired error rate when no deletion or updating occur. We then examine the effect of increasing deletion rate, i.e. the number of deleted messages divided by the total number of stored messages, for two values of $w_{MAX}$: 1 and 3, as shown in Fig. 3. It is observed that increasing the deletion rate when using binary connections in SCN will increase the MER especially after 4 iterations as the number of lost connections are increased. However, for deletion rates larger than 0.3, this increase in the MER is slowed down and the MER starts to decrease after deletion rates larger than 0.6. This is due to the fact that deletion of messages also reduces the density of the network as the cliques are removed. On the other hand, it can also be observed that using multiple-valued connection weights will resolve this problem as connections are never lost unless the number of cliques sharing a connection exceeds $w_{MAX}$. In that case, the error correcting capabilities of Architecture II will help recover the lost connections to an extent depending on the correlation of the cliques and the number of shared connections between them.

In another experiment, the effect of increasing the value of the integer weights were investigated on MER for all the three architectures as shown in Fig. 4. When $w_{MAX}$ is equal to 1, i.e., binary connections are used in storing the cliques. As $w_{MAX}$ is increased, the effect of the MER reduction in Architecture II is the best among the others although Architecture III also follows similar error rates to Architecture II after $w_{MAX}$ equals 4. Due to the presence of falsely activated nodes in Architecture I, its error performance is the worst among the others after $w_{MAX}$ equals 3. Therefore, if a low-complexity design is more important than reducing the error rate, Architecture III is the best option. On the other hand, if the error rate is a key design requirement, Architecture II is preferred.

The effect of deleting and adding new messages at equal rates (0.5) to the network is depicted in Fig. 5 by sweeping the value of $w_{MAX}$ and observing the behaviour or MER. Similar to the deletion-only scenario, Architecture II and III perform similarly for $w_{MAX} \geq 3$ while Architecture I again performs the worst among the others confirming the fact that it is essential to normalize the weights before feeding them to the global decoder.

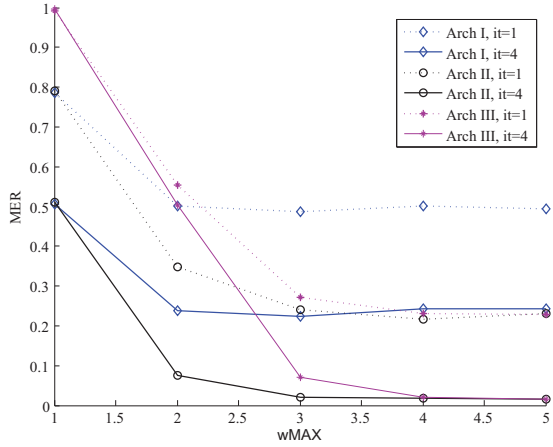

Fig. 4. The effect of increasing the maximum value of the connection weights on MER in the MV-SCN architectures ( $d = 0.4$, $ce = 0.5$, deletion rate = 0.5, addition rate = 0).

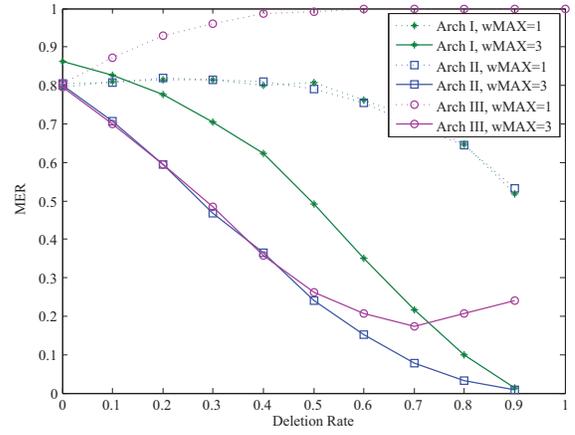

Fig. 6. Comparison of the MER for Architectures I, II, and III as the deletion rate is increased for both binary-weighted and Multiple-valued connections (it=1).

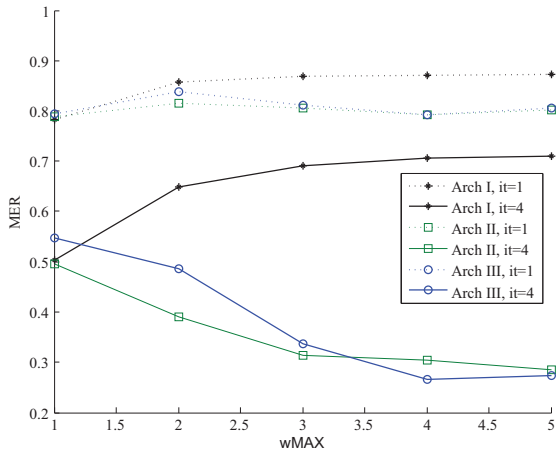

Fig. 5. The effect of increasing the maximum value of the connection weights on MER in the MV-SCN architectures ( $d = 0.4$, $ce = 0.5$, deletion rate = 0.5, addition rate = 0.5).

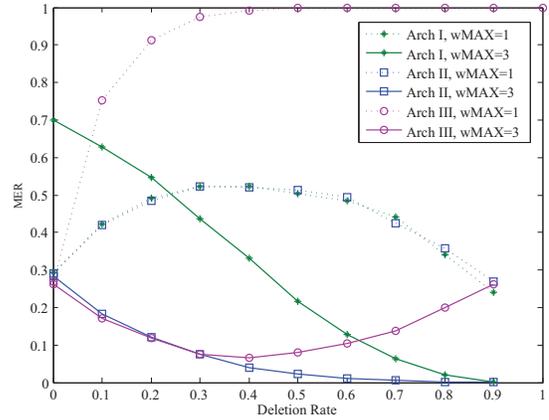

Fig. 7. Comparison of the MER for Architectures I, II, and III as the deletion rate is increased for both binary-weighted and Multiple-valued connections (it=4).

Fig. 6 and Fig. 7 show a comparison of the effect of the deletion rate on the MER for the three architectures after one and four iterations respectively. With the exception of Architecture I that has higher MERs than the others at zero deletion rate, the other two architectures have similar error rates. Architecture II and III have similar error rates up to deletion rates around 0.3 from where Architecture II performs better at higher MERs than Architecture III afterwards.

We have seen that using multiple-valued connections in SCN reduces the message error rate when deleting and updating messages to levels of MER comparable to that of the binary-weighted-based counterpart when no deletions or updates were performed. One of the drawbacks of incorporating multiple-valued connection into the SCN is that increasing the connection levels translates into the use of larger number of memory bits to store them. On the other hand, increasing the number of nodes in the networks introduces the opportunity to store the same number of messages but with lower density and thus lower MER results. We have also seen that choosing $w_{MAX}$ to be equal to 3, which results in doubling the required memory bits, shows promising results. To investigate the benefit of using multiple-valued connection weights, we experiment the same number of messages in a larger binary-weighted network such that the required memory usages would be equal. Therefore, doubling the memory size to store the multiple-valued weights is equivalent to increasing the number of nodes by $\sqrt{2}$ in a binary-weighted counterpart as the number of used memory bits is quadratically dependent on the number of nodes. [4]. A comparison of the error performance for both Architectures II and III is depicted in Fig. 8 and Fig. 9 respectively. As an example to realize the significance of the error rate reduction, the MER achieved from the Architecture II in our sample network, with 60% deleted contents and loaded with 40% density, is 12.8-fold smaller than that of the previous work with a similar amount of memory usage. In Architecture II, using the $w_{MAX}$ equal 3 reduces the MER more than the increased-size binary-weighted counterpart after deletion rates of around 0.2, while in Architecture III, the error rate significantly drops below the binary-weighted increased-size counterpart after deletion rates around 0.05 although the MER is increased again after deletion rates around 0.4 unlike Architecture II due to the error-correcting limitations of Architecture III.

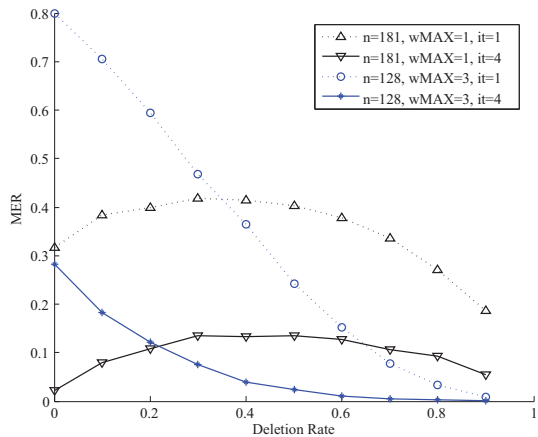

Fig. 8. The effect of increasing the deletion rate on the MER in Architecture II for a binary-weighted SCN with double the number of connections storing the same number of messages as the multiple-valued weight-coded SCN.

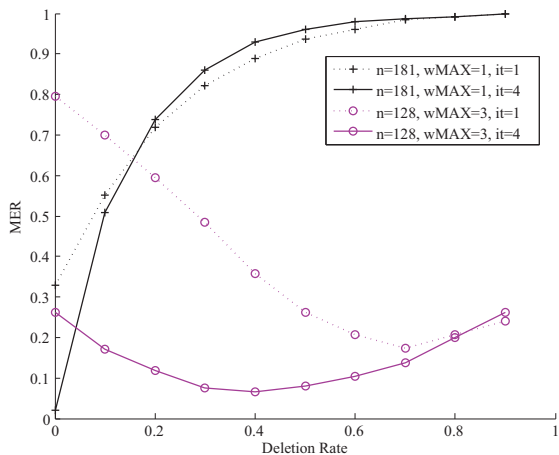

Fig. 9. The effect of increasing the deletion rate on the MER in Architecture III for a binary-weighted SCN with double the number of connections storing the same number of messages as the multiple-valued weight-coded SCN.

## V. CONCLUSION

In this paper, we addressed an important drawback with binary-weighted associative memories based on sparse clustered networks when deleting or updating the stored messages. As the stored patterns share data due to the way they are represented in the network, i.e. binary connections between the nodes, removing a message may result in removing a connection or connections that are used for other stored messages that are not intended to be deleted or updated. This problem is exacerbated when the density of the network, which is directly proportional to the number of different messages the network can store, is increased. We have shown that updating messages can recover some of the removed connections from the remaining patterns. We proposed an algorithm that incorporates multiple-valued weighted-connections into the associative memory instead of the binary ones, and showed that the proposed method significantly improves the error rate in retrieval process of the stored messages. Two different architectures based on variations of previous work were proposed and investigated: First an architecture that can correct erroneous inputs as well as partial ones with the cost of hardware resources. Second, a reduced-complexity architecture that can only be used in applications with partial input patterns.

REFERENCES

[1] J. J. Hopfield, "Neural networks and physical systems with emergent collective computational abilities," *Proceedings of the National Academy of Sciences of the USA*, vol. 79, no. 8, pp. 2554–2558, Apr. 1982.
[2] V. Gripon and C. Berrou, "Sparse neural networks with large learning diversity," *IEEE Transactions on Neural Networks*, vol. 22, no. 7, pp. 1087–1096, Jul. 2011.
[3] H. Jarollahi, N. Onizawa, V. Gripon, and W. J. Gross, "Architecture and implementation of an associative memory using sparse clustered networks," in *IEEE International Symposium on Circuits and Systems (ISCAS)*, Seoul, Korea, May 2012, pp. 2901–2904.
[4] ——, "Reduced-complexity binary-weight-coded associative memories," in *Proceedings of the 2013 IEEE International Conference on Acoustics, Speech, and Signal Processing (ICASSP)*, May 2013, pp. 2523–2527.
[5] O. Qadir, J. Liu, J. Timmis, G. Tempesti, and A. Tyrrell, "Hardware architecture for a bidirectional hetero-associative protein processing associative memory," in *2011 IEEE Congress on Evolutionary Computation (CEC)*, Jun. 2011, pp. 208–215.
[6] V. Gripon and C. Berrou, "Nearly-optimal associative memories based on distributed constant weight codes," in *Information Theory and Applications Workshop (ITA), 2012*, Feb. 2012, pp. 269–273.
[7] H. Jarollahi, N. Onizawa, and W. Gross, "Selective decoding in associative memories based on sparse-clustered networks," in *Proceedings of 2013 IEEE Global Conference on Signal and Information Processing*, Dec. 2013, to Appear.